\documentclass[runningheads]{llncs}

\usepackage[T1]{fontenc}
\usepackage{graphicx}
\graphicspath{{./}{fig/}{figures/}{../fig/}{../figures/}}
\usepackage{amsmath,amssymb}
\usepackage{booktabs}
\usepackage{multirow}
\usepackage{url}
\usepackage{microtype}
\usepackage{placeins}

\title{DAEP: Difficulty-Aware Evidence Planning for Medical Video Corpus Temporal Answer Grounding}
\titlerunning{Difficulty-Aware Evidence Planning}

\author{Tianjian He\inst{1} \and
Yujie Liu\inst{2} \and
Zhiping Huang\inst{3} \and
Changbo Xu\inst{2}\thanks{Corresponding author.}}
\authorrunning{T. He et al.}
\institute{TikTok, ByteDance, China \and
Beijing Institute of Graphic Communication, Beijing, China\\
\email{20240353011z@stu.bigc.edu.cn}\\
\email{xuchangbo@bigc.edu.cn} \and
Lingnan University, Hong Kong, China\\
\email{zhipinghuang@ln.hk}}

\begin{document}
\maketitle

\begin{abstract}
We present DAEP, team BIGC's system for Track 3 of the NLPCC 2026 Shared Task 1: Difficulty-Aware Temporal Answer Grounding in Video Corpus (DA-TAGVC). The task requires retrieving a target video from 50 candidates and localizing the answer-supporting segment. DAEP ranks candidates with subtitle, visual, and procedural-context evidence before anchor expansion and span reranking. The task-provided Simple/Complex label drives an evidence plan that controls modality weights, Top-\(K\) aggregation, boundary expansion, and reranking. BIGC ranked first among ten systems on all four official metrics, achieving an Average score of 0.2728 and a 9.4\% relative improvement over the runner-up. Validation Average falls from 0.2646 to 0.2245 without the planner; further ablations support the contributions of visual evidence, context, and span reranking.

\end{abstract}

\section{Introduction}

Medical instructional videos combine visual demonstrations, subtitles, and procedural context; many medical questions therefore require a supporting video segment rather than a text answer alone~\cite{gupta2023medvidqa}. We describe the BIGC submission to DA-MIVQA, the NLPCC 2026 Shared Task 1 on difficulty-aware multilingual and multimodal medical instructional video understanding~\cite{nlpcc2026sharedtasks}. We focus on Track 3, DA-TAGVC, where a system receives a question and 50 candidate videos, retrieves the target video, and returns ranked answer spans.

\begin{figure}[!htbp]
\centering
\includegraphics[width=\linewidth]{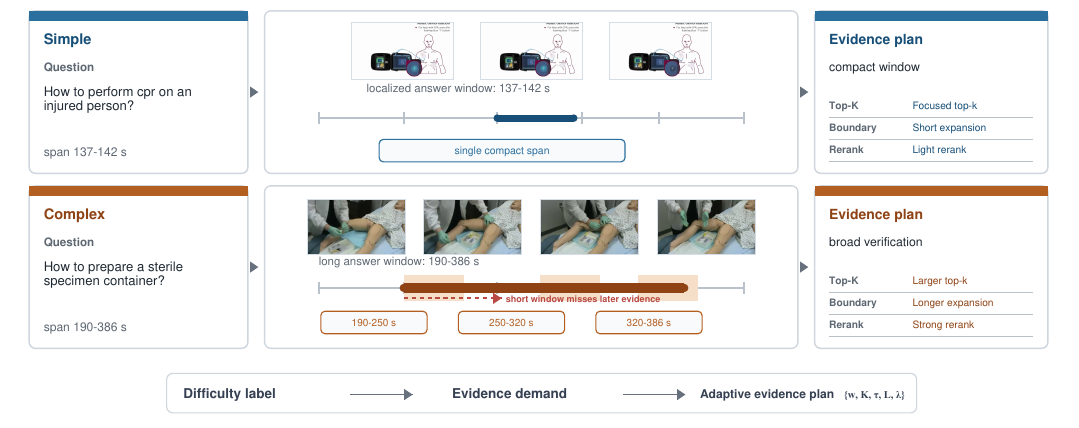}
\caption{Motivation for difficulty-aware evidence planning, illustrated with released Track 3 training examples. The examples show how the supplied label can lead the planner to consider different evidence scopes; they are not a distributional comparison of answer duration.}
\label{fig:motivation}
\end{figure}
\FloatBarrier

DA-TAGVC couples video retrieval with temporal boundary prediction, extending the NLPCC medical instructional video QA series~\cite{li2023cmivqa,li2025m4ivqa,liu2025m3med}. Questions with different difficulty labels may require different combinations of local subtitle evidence, surrounding context, and cross-modal verification. Figure~\ref{fig:motivation} illustrates two such cases with released training examples. DAEP retains the established VCMR retrieval-localization decomposition~\cite{paul2021hman,zhang2021reloclnet,hou2021conquer,lei2024prem}, but treats the official difficulty label as an evidence-allocation signal that coordinates modality weighting, retrieval breadth, boundary expansion, and span verification. We report the official leaderboard, stratified validation results, component ablations, planner-control comparisons, and their evidential scope.

\section{Related Work}

Medical instructional video QA often requires localized visual support beyond text answers~\cite{gupta2023medvidqa}. NLPCC's CMIVQA, MMIVQA, M4IVQA, and M3-Med tasks progressively add corpus retrieval, multilingual questions, multimodal reasoning, and multi-hop evidence~\cite{li2023cmivqa,li2025mmivqa,li2025m4ivqa,liu2025m3med}. Grounded video QA likewise benefits from locating answer evidence and transferring knowledge across modalities~\cite{lei2020tvqa,li2023learning,weng2023mutual,li2024visualprompt}. The 2026 overview introduces DA-TAGVC's simple/complex grounding protocol~\cite{nlpcc2026sharedtasks}.

Temporal grounding has developed from language-based moment localization to span, dense-regression, and transformer decoders~\cite{gao2017tall,hendricks2017localizing,zhang2020vslnet,zeng2020drn,lei2021momentdetr,lin2023univtg}. Video corpus moment retrieval then adds corpus-level ranking before localization: TVR formalizes video-subtitle retrieval~\cite{lei2020tvr}, while HMAN, ReLoCLNet, CONQUER, and PreM address retrieval, localization, query-aware reranking, or partial relevance~\cite{paul2021hman,zhang2021reloclnet,hou2021conquer,lei2024prem}. CLIP and BERT provide reusable visual-language and text encoders~\cite{radford2021clip,devlin2019bert}. Difficulty-aware learning has also been studied through curriculum learning and VideoQA difficulty analysis~\cite{bengio2009curriculum,terao2020difficulty,eyzaguirre2025complexity}. DAEP instead uses the task-provided simple/complex label as an inference-time interface that coordinates retrieval, evidence aggregation, boundary decoding, and reranking.

\FloatBarrier
\section{Method}

\begin{figure}[!t]
\centering
\includegraphics[width=\linewidth,trim=8pt 10pt 8pt 7pt,clip]{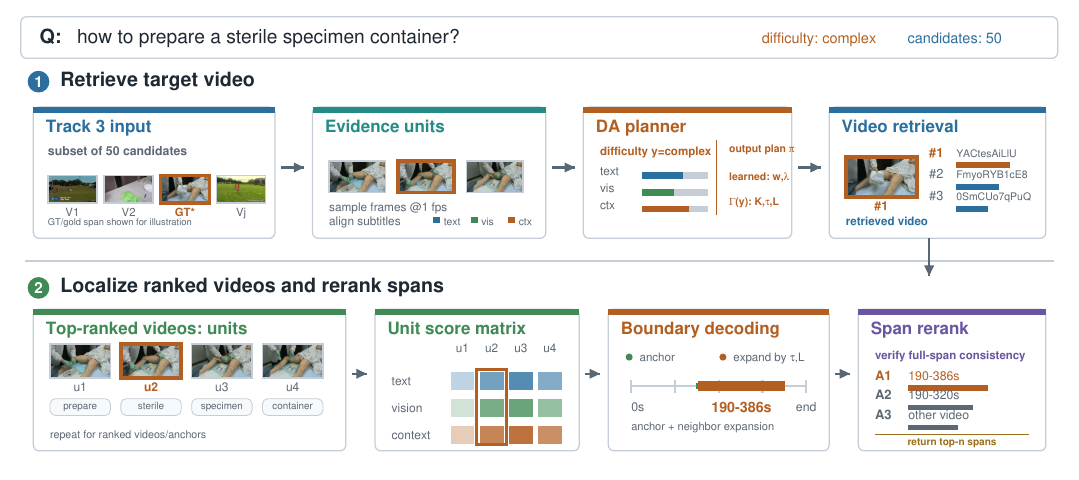}
\caption{DAEP overview on a released Track 3 training example, from candidate-video retrieval to temporal grounding, span reranking, and top-\(n\) output. The ground-truth video and gold span are shown only for illustration and are unavailable at validation and test time.}
\label{fig:method}
\end{figure}

Figure~\ref{fig:method} summarizes DAEP using a released Track 3 training example. Following the VCMR decomposition of retrieval and moment localization~\cite{paul2021hman,zhang2021reloclnet,hou2021conquer,lei2024prem}, DAEP encodes subtitle-aligned text, visual, and procedural-context evidence. Its hybrid plan $\pi_i=\{\mathbf{w}_i,K_i,\tau_i,L_i,\lambda_i\}$ controls video ranking, boundary decoding, and reranking. The figure follows one video path, while the system expands multiple anchors across ranked videos. The gold video and span are illustrative; validation and test use only \((q_i,\mathcal{V}_i,y_i)\).

\subsection{Task Formulation}

Each example contains a question $q_i$, 50 candidate videos $\mathcal{V}_i$, a difficulty label $y_i$, and training/validation annotations $(V_i^*,t_i^s,t_i^e)$. At inference, DAEP observes only $(q_i,\mathcal{V}_i,y_i)$ and returns ranked video-span candidates for R@n|mIoU with \(n\in\{1,10,100\}\). It constructs a hybrid evidence plan \(\pi_i=\{\mathbf{w}_i,K_i,\tau_i,L_i,\lambda_i\}\): \(\mathbf{w}_i\) and \(\lambda_i\) are learned per question, whereas \(K_i,\tau_i,L_i\) come from a difficulty-conditioned lookup fixed after validation.

\subsection{Subtitle-Aligned Evidence Unit Construction}

Following video-subtitle retrieval settings such as TVR~\cite{lei2020tvr}, DAEP represents each candidate video as timestamped subtitle-aligned evidence units \(V_{i,j}=\{s_{i,j,k}\}_{k=1}^{M_j}\), where \(s_{i,j,k}=(x_{i,j,k},f_{i,j,k},c_{i,j,k},t_{i,j,k}^s,t_{i,j,k}^e)\) stores subtitle text, sampled frames, context, and timestamps.

\subsection{Feature Encoding}

Question, subtitle, visual, and context signals are encoded for each unit. The question $q_i$ and subtitle text $x_{i,j,k}$ use BERT-style Transformer representations~\cite{devlin2019bert}. Visual evidence is sampled at 1 fps, encoded with CLIP ViT-B/32~\cite{radford2021clip}, and mean-pooled within the unit. The context $c_{i,j,k}$ concatenates the current subtitle, up to two neighboring units on each side, and question-matched procedural cues.

Let $u_q,u^t,u^v,u^c$ denote the raw representations. Modality-specific projection heads $P_q,P_t,P_v,P_c$ map them into a shared $d$-dimensional matching space, followed by L2 normalization: $\mathbf{h}_{q_i}=\operatorname{norm}(P_q u_q)$ and $z^m_{i,j,k}=\operatorname{norm}(P_m u^m_{i,j,k})$ for $m\in\{t,v,c\}$. Matching uses cosine similarity between normalized question and unit features.

\subsection{Procedural Context Evidence}

Procedural context enriches evidence rather than generating an answer. Each context concatenates the current subtitle with up to two units on either side. DAEP also matches salient procedural terms from the question to subtitle units and their local neighborhoods using lexical and embedding similarity, then appends the matched cues before projection. The same multilingual text encoder represents Chinese and English. Cue extraction uses only candidate content, does not change the candidate set, and accesses no target-video or span labels~\cite{hou2021conquer,lei2024prem}.

\subsection{Difficulty-Aware Evidence Planner}

DAEP first computes initial modality matches \(r_i^m=\max_{j,k}\cos(\mathbf{h}_{q_i},z^m_{i,j,k})\), \(m\in\{t,v,c\}\). Its hybrid planner learns question-specific modality weights and reranking strength; discrete candidate controls are calibrated by difficulty group on validation data:
\begin{equation}
\begin{aligned}
\mathbf{u}_i &= \operatorname{MLP}([\mathbf{h}_{q_i};y_i;r_i^t;r_i^v;r_i^c]),\\
\mathbf{w}_i &= \operatorname{softmax}(W_w\mathbf{u}_i+b_w),\qquad
\lambda_i=\lambda_{\max}\sigma(W_\lambda\mathbf{u}_i+b_\lambda),\\
(K_i,\tau_i,L_i) &= \Gamma(y_i),\qquad
\pi_i=\{\mathbf{w}_i,K_i,\tau_i,L_i,\lambda_i\}.
\end{aligned}
\end{equation}
Here \(\Gamma\) is selected separately for Simple and Complex questions from the bounded \((K,\tau,L)\) grid described in Sect.~4.2 and fixed thereafter. Gradients update the encoders, projections, \(\mathbf{w}_i\), and \(\lambda_i\), but not the discrete lookup. The resulting planner is an inspectable conditional search rather than a learned continuous mapping from difficulty to the three controls.

\subsection{Stage 1: Candidate Video Retrieval}

For each evidence unit $s$, DAEP computes modality-specific cosine scores and fuses them as \(S(q_i,s)=w_i^tS_t(q_i,s)+w_i^vS_v(q_i,s)+w_i^cS_c(q_i,s)\). The video-level score uses adaptive Top-\(K_i\) aggregation, following VCMR practice of pooling query-relevant unit evidence into video scores~\cite{zhang2021reloclnet,hou2021conquer,lei2024prem}:
\begin{equation}
\begin{aligned}
S(q_i,V_{i,j}) &=
\frac{1}{K_i}\sum_{s\in\mathrm{TopK}_i(V_{i,j})} S(q_i,s),\\
\mathcal{R}_i^{\downarrow} &= \operatorname{sort}_{V_{i,j}\in\mathcal{V}_i} S(q_i,V_{i,j}),\\
\hat{V}_i &= \mathcal{R}_i^{\downarrow}[1].
\end{aligned}
\end{equation}
Here $\mathrm{TopK}_i(V_{i,j})$ returns the $K_i$ highest-scoring units, and $\mathcal{R}_i^{\downarrow}$ is the ranked video list.

\subsection{Stage 2: Temporal Boundary Decoding}

DAEP decodes candidate spans from ranked videos and then reranks them. Local NMS proceeds greedily in descending unit-score order and suppresses units within \(D_{\mathrm{nms}}\) subtitle positions of an accepted anchor. The top \(H\) remaining units in each candidate video are selected as anchors:
\begin{equation}
\begin{aligned}
\{s_{i,j,h}^*\}_{h=1}^{H} &= \operatorname{TopH}_{\mathrm{nms}}\{S(q_i,s)\mid s\in V_{i,j}\},\\
C_{i,j,h} &= \{s \mid \operatorname{Adj}(s,s_{i,j,h}^*)=1,\ S(q_i,s)\geq\tau_i,\ \operatorname{Dist}(s,s_{i,j,h}^*)\leq L_i\},\\
\hat{t}_{i,j,h}^s &= \min_{s\in C_{i,j,h}\cup\{s_{i,j,h}^*\}} t_s(s), \\
\hat{t}_{i,j,h}^e &= \max_{s\in C_{i,j,h}\cup\{s_{i,j,h}^*\}} t_e(s),\\
a_{i,j,h} &= (V_{i,j},\hat{t}_{i,j,h}^s,\hat{t}_{i,j,h}^e).
\end{aligned}
\end{equation}
\(\operatorname{Adj}\) and \(\operatorname{Dist}\) encode temporal adjacency and unit distance. The pool \(\tilde{\mathcal{A}}_i=\{a_{i,j,h}\mid V_{i,j}\in\mathcal{R}_i^{\downarrow}[1:M_v],1\leq h\leq H\}\) is reduced to the top \(M\) spans by base score.

\subsection{Span Consistency Reranking}

Given $\mathcal{A}_i=\{a_{i,1},\ldots,a_{i,M}\}$, DAEP scores all spans for the final ranked list. Let $V(a)$ be the source video and $\mathcal{S}(a)$ the units covered by span $a$. The retrieval-aware base score and span-consistency score are
\begin{equation}
\begin{aligned}
B(a_{i,m}) &=
S(q_i,V(a_{i,m}))+
\frac{1}{|\mathcal{S}(a_{i,m})|}
\sum_{s\in\mathcal{S}(a_{i,m})} S(q_i,s),\\
R(a_{i,m}) &= w_i^tC^t(a_{i,m})+w_i^vC^v(a_{i,m})+w_i^cC^c(a_{i,m}).
\end{aligned}
\end{equation}
Here \(C^t,C^v,C^c\) are cosine similarities between the question and span-level subtitle, frame, and context encodings. DAEP scores each span by
\begin{equation}
\operatorname{Score}(a_{i,m})=
B(a_{i,m})+\lambda_iR(a_{i,m}),
\end{equation}
and returns the top-\(n\) spans. This reranker scores each decoded span as a whole, following query-aware VCMR reranking while conditioning its strength on task difficulty~\cite{hou2021conquer,lei2024prem}.

\subsection{Training Objective}

Training optimizes video ranking and span scoring over generated candidates. DAEP back-propagates through the trainable text encoder, projections, modality weights, reranking strength, and span scores, but not through the frozen visual encoder, Top-\(K_i\), Top-\(H\) anchor/NMS selection, thresholding, or neighborhood expansion.

Let \(g_i=(V_i^*,t_i^s,t_i^e)\), and let \(a_i^+\) be the target-video candidate with the highest IoU. Training minimizes \(\mathcal{L}_{rank}+\eta_1\mathcal{L}_{span}+\eta_2\mathcal{L}_{iou}\). The first term is video-level cross entropy. Candidate probabilities are \(p_i(a)=\operatorname{softmax}_{a\in\mathcal{A}_i}(\operatorname{Score}(a))\), giving \(\mathcal{L}_{span}=-\log p_i(a_i^+)\) and \(\mathcal{L}_{iou}=\sum_a p_i(a)(1-\operatorname{IoU}(a,g_i))\). If no generated span has positive IoU, both span losses are masked and only \(\mathcal{L}_{rank}\) is applied. The IoU term therefore trains candidate scores, not hard boundary selection.

\section{Experiments}

\subsection{Dataset and Evaluation}

We evaluate DAEP on DA-TAGVC, the Track 3 subset of NLPCC 2026 Shared Task 1~\cite{nlpcc2026sharedtasks}. Each question has 50 candidate videos and a binary difficulty label. Table~\ref{tab:data_stats} reports train/validation statistics, showing that difficulty is not simply answer duration. The hidden-test input contains 394 Chinese-Simple, 280 Chinese-Complex, 210 English-Simple, and 234 English-Complex questions; target videos and answer spans are not included in the participant-facing test input files.

\begin{table}[!ht]
\centering
\small
\caption{Training and validation-set statistics of the DA-TAGVC dataset.}
\label{tab:data_stats}
\begin{tabular}{llrrrr}
\toprule
Split & Group & \#Q & \#Targets & Cand./Q & Avg. Span \\
\midrule
Train & Chinese-Simple & 2101 & 930 & 50 & 37.70s \\
Train & Chinese-Complex & 1842 & 826 & 50 & 34.72s \\
Train & English-Simple & 1273 & 556 & 50 & 72.87s \\
Train & English-Complex & 1341 & 480 & 50 & 45.01s \\
Val & Chinese-Simple & 362 & 155 & 50 & 41.10s \\
Val & Chinese-Complex & 285 & 137 & 50 & 33.34s \\
Val & English-Simple & 199 & 92 & 50 & 69.30s \\
Val & English-Complex & 223 & 80 & 50 & 43.27s \\
\bottomrule
\end{tabular}
\end{table}

The official Track 3 metrics are R@1|mIoU, R@10|mIoU, R@100|mIoU, and Average~\cite{nlpcc2026sharedtasks}. They measure temporal localization quality at retrieval depths 1, 10, and 100; Average is their arithmetic mean. We follow the released evaluation protocol and output format, assigning zero temporal overlap to a prediction from a non-target video. DAEP retains up to 100 ranked video-span candidates by expanding multiple anchors in the retrieved video list. Official comparisons use the released leaderboard; validation ablations and sanity checks use the same split, candidate sets, and evaluator.

\subsection{Implementation Details}

Questions, subtitles, and context use multilingual BERT (bert-base-multilingual-cased)~\cite{devlin2019bert}, updated during training. Visual features use frozen CLIP ViT-B/32 (openai/clip-vit-base-patch32)~\cite{radford2021clip} and are cached after 1-fps sampling. The matching dimension is \(d=256\); projection heads are Linear--GELU--Dropout--Linear, and the two-layer planner MLP uses the same hidden size. The validation grid contains 80 tuples per difficulty group: $K\in\{1,\ldots,5\}$, $\tau\in\{0.35,0.45,0.55,0.65\}$, and $L\in\{0,\ldots,3\}$. The test lookup is \(\Gamma(\mathrm{Simple})=(2,0.55,1)\) and \(\Gamma(\mathrm{Complex})=(5,0.35,3)\). We use $D_{\mathrm{nms}}=2$, $\lambda_{\max}=0.30$, $M_v=50$, $H=2$, $M=100$, and $\eta_1=\eta_2=1$; hence, at most 100 spans are decoded before reranking. Training uses AdamW for 15 epochs, batch size 16, 10\% warmup, and early stopping on validation Average. Learning rates are $2\times10^{-5}$ for mBERT and $1\times10^{-4}$ for trainable heads. The configuration is Python 3.10, PyTorch 2.2, Transformers 4.41, CUDA 12.1, one NVIDIA A100-40GB GPU, and seed 2026; training takes about 5.2 hours with cached visual features. The official submission uses this single validation-selected run. Repeated-seed variance and end-to-end inference latency are not reported.

\subsection{Resource and Rule Compliance}

Training uses the official split, model selection uses validation, and test predictions use the selected checkpoint. The provided difficulty label is used directly, never inferred from spans or hidden labels. External weights are limited to the generic encoders in Sect.~4.2: mBERT is fine-tuned on official training data and CLIP is frozen. We use no external medical annotations, video-QA data, commercial medical models, or additional medical pretraining. All remaining components use official train and validation data. Figures~\ref{fig:motivation} and~\ref{fig:method} use released training examples rather than hidden test material.

\FloatBarrier
\subsection{Official Evaluation Results}

Table~\ref{tab:leaderboard} reproduces the official Track 3 leaderboard~\cite{nlpcc2026sharedtasks}; citations are retained where the overview links a prior report or organizer baseline. BIGC ranks first among ten systems on all four metrics. Because participant-facing test files omit gold videos and spans and no official subgroup scores were released, the language- and difficulty-stratified results below use validation annotations.

\begin{table}[!ht]
\centering
\small
\caption{Official Track 3 leaderboard~\cite{nlpcc2026sharedtasks}. Metrics follow protocol order.}
\label{tab:leaderboard}
\begin{tabular}{clcccc}
\toprule
Rank & Team ID & R@1|mIoU & R@10|mIoU & R@100|mIoU & Average \\
\midrule
\textbf{1} & \textbf{BIGC} & \textbf{0.1646} & \textbf{0.2879} & \textbf{0.3660} & \textbf{0.2728} \\
2 & UWM & 0.1415 & 0.2727 & 0.3338 & 0.2493 \\
3 & IIEleven~\cite{ma2024multilingual} & 0.1238 & 0.2513 & 0.3566 & 0.2439 \\
4 & UESC & 0.1364 & 0.2551 & 0.3359 & 0.2425 \\
5 & MedEcho~\cite{zhou2025multihop} & 0.1490 & 0.2588 & 0.3152 & 0.2410 \\
6 & Nsddd~\cite{cheng2023unified} & 0.1436 & 0.2129 & 0.3235 & 0.2267 \\
7 & DesiWen & 0.1232 & 0.2351 & 0.3159 & 0.2248 \\
8 & MM-Baseline~\cite{weng2023mutual} & 0.1028 & 0.2572 & 0.3082 & 0.2228 \\
9 & Text-Only Baseline & 0.0885 & 0.2415 & 0.2853 & 0.2051 \\
10 & Random Pick Method~\cite{li2023cmivqa} & 0.0454 & 0.0872 & 0.0673 & 0.0666 \\
\bottomrule
\end{tabular}
\end{table}

\FloatBarrier
\subsection{Validation Analysis}

Because DA-TAGVC is difficulty-aware, Table~\ref{tab:difficulty_split} reports validation results by language and difficulty. DAEP reaches an Average score of 0.2729 on Simple questions and 0.2554 on Complex questions. Without the planner, these scores fall to 0.2400 and 0.2075, respectively. The larger reduction for Complex questions supports difficulty-conditioned planning. English-Simple questions also score below Chinese-Simple questions and have longer mean answer spans (Table~\ref{tab:data_stats}); this association is consistent with greater boundary difficulty but does not establish causality.

\begin{table}[!ht]
\centering
\small
\setlength{\tabcolsep}{3pt}
\caption{Validation results by language and difficulty group. R@n denotes R@n|mIoU.}
\label{tab:difficulty_split}
\begin{tabular}{llrrrrr}
\toprule
Method & Group & \#Q & R@1 & R@10 & R@100 & Avg. \\
\midrule
DAEP & Chinese-Simple & 362 & 0.1734 & 0.2920 & 0.3625 & 0.2760 \\
DAEP & Chinese-Complex & 285 & 0.1500 & 0.2738 & 0.3516 & 0.2585 \\
DAEP & English-Simple & 199 & 0.1632 & 0.2824 & 0.3565 & 0.2674 \\
DAEP & English-Complex & 223 & 0.1414 & 0.2654 & 0.3476 & 0.2515 \\
w/o DA Planner & Chinese-Simple & 362 & 0.1460 & 0.2551 & 0.3268 & 0.2426 \\
w/o DA Planner & Chinese-Complex & 285 & 0.1160 & 0.2210 & 0.2945 & 0.2105 \\
w/o DA Planner & English-Simple & 199 & 0.1381 & 0.2464 & 0.3211 & 0.2352 \\
w/o DA Planner & English-Complex & 223 & 0.1085 & 0.2130 & 0.2892 & 0.2036 \\
\bottomrule
\end{tabular}
\end{table}

Table~\ref{tab:ablation} reports DAEP-internal ablations under identical splits, encoders, optimizer, candidate sets, and evaluation code. Each variant disables one component in Fig.~\ref{fig:method}. Removing visual evidence, context evidence, span reranking, and the planner reduces Average by 0.0170, 0.0254, 0.0097, and 0.0401, respectively. The planner ablation uses uniform modality weights, \(K=3\), \(\tau=0.45\), \(L=2\), and \(\lambda=0.15\).

\begin{table}[!ht]
\centering
\small
\setlength{\tabcolsep}{4pt}
\caption{Validation-set component ablations. All rows are DAEP variants; DA denotes difficulty-aware.}
\label{tab:ablation}
\begin{tabular}{lcccc}
\toprule
Method & R@1|mIoU & R@10|mIoU & R@100|mIoU & Average \\
\midrule
\textbf{DAEP (ours)} & \textbf{0.1586} & \textbf{0.2798} & \textbf{0.3554} & \textbf{0.2646} \\
w/o Visual Evidence & 0.1449 & 0.2626 & 0.3352 & 0.2476 \\
w/o Context Evidence & 0.1392 & 0.2524 & 0.3261 & 0.2392 \\
w/o Span Reranking & 0.1513 & 0.2691 & 0.3442 & 0.2549 \\
w/o DA Planner & 0.1287 & 0.2356 & 0.3093 & 0.2245 \\
\bottomrule
\end{tabular}
\end{table}

\begin{table}[!ht]
\centering
\small
\setlength{\tabcolsep}{6pt}
\caption{Planner-control validation results. Values are Average scores; aggregates are weighted by validation group size before rounding.}
\label{tab:planner_control}
\begin{tabular}{lccc}
\toprule
Method & Simple & Complex & All \\
\midrule
\textbf{DAEP} & \textbf{0.2729} & \textbf{0.2554} & \textbf{0.2646} \\
Fixed difficulty & 0.2528 & 0.2288 & 0.2414 \\
Shuffled difficulty & 0.2477 & 0.2163 & 0.2328 \\
w/o DA Planner & 0.2400 & 0.2075 & 0.2245 \\
\bottomrule
\end{tabular}
\end{table}

\FloatBarrier
Table~\ref{tab:planner_control} examines whether the label-conditioned controls carry information beyond a fixed setting. Fixed difficulty assigns every question to the stronger Simple branch. Shuffled difficulty uses one fixed permutation that preserves label counts but breaks question-label alignment. Both retain the learned heads and underperform DAEP, with larger gaps on Complex questions. The order prior keeps the supplied candidate order and uses the language--difficulty mean training span. The frequency prior ranks candidates by training target frequency and uses a video-specific mean training span when available, otherwise the group mean. They score 0.0282 and 0.0603 Average, respectively. Because the lookup and checkpoint are selected on validation, these comparisons are diagnostic rather than unbiased test estimates.

BIGC exceeds the best non-BIGC score in each leaderboard column by 0.0156 on R@1|mIoU, 0.0152 on R@10|mIoU, 0.0094 on R@100|mIoU, and 0.0235 on Average. The Average difference corresponds to a 9.4\% relative improvement. The leaderboard establishes the competitiveness of the complete submitted system, whereas Tables~\ref{tab:ablation} and~\ref{tab:planner_control} provide controlled validation comparisons within DAEP. Systems on the leaderboard remain independent implementations rather than shared-backbone baselines.

Manual inspection of validation predictions revealed examples of four qualitative failure modes: confusion between similar procedures, drift between narration and action boundaries, subtitle--visual mismatch, and inconsistent bilingual terminology. These observations are not error-frequency estimates. The official Average score of 0.2728 leaves substantial room for improvement; stage-specific metrics would be needed to separate retrieval from boundary errors.

\clearpage
\section{Discussion}

Retrieval-localization is established in VCMR~\cite{paul2021hman,zhang2021reloclnet,hou2021conquer,lei2024prem}, and difficulty modeling is established in adaptive learning~\cite{bengio2009curriculum,eyzaguirre2025complexity}. DAEP builds on these designs. Its contribution is an explicit interface between the supplied label and four evidence decisions: modality emphasis, aggregation, boundary expansion, and span reranking.

DAEP assumes the benchmark-supplied Simple/Complex label. Applications without it would require estimated evidence demand or a label-agnostic fallback, neither evaluated here. The study uses one validation-selected run and reports no repeated-seed variance, confidence intervals, significance tests, alternative validation splits, or grid sensitivity. Positive-IoU candidate coverage and end-to-end latency are also not reported. These limits constrain robustness and deployment claims, not the reported official ranking or within-run validation comparisons.

The joint metric cannot identify whether retrieval or localization caused an error. Future work can report video recall and conditional span IoU, estimate continuous difficulty, learn boundary refinement, quantify error types, and add shared-backbone baselines. DAEP is intended to retrieve evidence for human review, not to make clinical decisions.

\section{Conclusion}

We presented DAEP, team BIGC's system for NLPCC 2026 Shared Task 1 Track 3 DA-TAGVC. DAEP ranks candidate videos, decodes temporal spans, and uses the task-provided Simple/Complex label to coordinate evidence weighting, retrieval breadth, boundary expansion, and span reranking. BIGC ranked first among ten systems on every official metric and obtained an Average score of 0.2728. Validation ablations further show that multimodal evidence, span reranking, and difficulty-conditioned planning improve the complete system under the reported validation setting. These results establish DAEP as an effective shared-task system and motivate future work on predicted difficulty and finer boundary modeling.

\begin{credits}
\subsubsection{\discintname}
The authors have no competing interests to declare that are relevant to the content of this article.
\end{credits}

\bibliographystyle{splncs04}
\bibliography{reference}

\end{document}